\documentclass[10pt,twocolumn,letterpaper]{article}

\usepackage{cvpr}
\usepackage{times}
\usepackage{epsfig}
\usepackage{graphicx}
\usepackage{amsmath}
\usepackage{amssymb}

\usepackage{amssymb}
\usepackage{enumitem}
\usepackage{lettrine}
\usepackage{flushend}
\usepackage[super]{nth}
\usepackage{xspace}

\usepackage[pagebackref=true,breaklinks=true,letterpaper=true,colorlinks,bookmarks=false]{hyperref}

\cvprfinalcopy

\begin{document}

%%%%%%%%% TITLE
\title{Multiple Measurements and Joint Dimensionality Reduction for Large Scale Image Search with Short Vectors\\
\vspace{5mm} \large Extended Version}

\author{Filip Radenovi\'{c}$^1$ \quad Herv\'{e} J\'{e}gou$^2$ \quad Ond\v{r}ej Chum$^1$\\
$^1$CMP, Faculty of Electrical Engineering, Czech Technical University in Prague\\
$^2$INRIA, Rennes\\
{\tt\small filip.radenovic@cmp.felk.cvut.cz, herve.jegou@inria.fr, chum@cmp.felk.cvut.cz}
}

\maketitle

%%%%%%%%% ABSTRACT
\begin{abstract}

This paper addresses the construction of a short-vector\linebreak (128D) image representation for large-scale image and particular object retrieval. In particular, the method of joint dimensionality reduction of multiple vocabularies is considered.
We study a variety of vocabulary generation techniques: different k-means initializations, different descriptor transformations, different measurement regions for descriptor extraction. Our extensive evaluation shows that different combinations of vocabularies, each partitioning the descriptor space in a different yet complementary manner, results in a significant performance improvement, which exceeds the state-of-the-art.

\end{abstract}

%%%%%%%%% BODY TEXT

\section{Introduction}

\lettrine{L}{arge-scale} image retrieval techniques have been developing and improving greatly for more than a decade. Many of the current state-of-the-art approaches~\cite{Nister-CVPR06,Chum-ICCV07,Jegou-IJCV10} are based on the bag-of-words (BOW) approach originally proposed by Sivic and Zisserman~\cite{Sivic-ICCV03}. Another popular image representation arises from aggregating local descriptors like Fisher kernel~\cite{Perronnin-CVPR10} and Vector of Locally Aggregated Descriptors (VLAD)~\cite{Jegou-CVPR10}.
%are developed as a variation of the original approach, but their representations differ significantly from the BOW representation.

The BOW vectors are high dimensional (up to 64 million dimensions in \cite{Mikulik-IJCV12}), so, due to the high memory and computational requirements, search is limited to a several million images on a single machine. There are more scalable approaches that tackle this problem by generating compact image representations~\cite{Torralba-CVPR08,Perronnin-CVPR10,Jegou-CVPR10}, where the image is described by a short vector that can be additionally compressed into compact codes using binarization~\cite{Torralba-CVPR08,Weiss-NIPS09}, product quantization~\cite{Jegou-PAMI11}, or recently proposed additive quantization techniques~\cite{Babenko-CVPR14}.
In this paper we propose and experimentally evaluate simple techniques that additionally boost retrieval performance, but at the same time preserve low memory and computational costs.

Short vector image representations are often generated using the principal component analysis (PCA)~\cite{Bishop06} technique to perform the dimensionality reduction over high-dimensional vectors. Jegou and Chum~\cite{Jegou-ECCV12} study the effects of PCA on BOW representations. They show that both steps of PCA procedure, i.e., centering and selection of de-correlated (orthogonal) basis minimizing the dimensionality reduction error, improve retieval performance. Centering (mean subtraction) of BOW vectors provides a boost in performance by adding a higher value to the negative evidence: given two BOW vectors, a visual word jointly missing in both vectors provides useful information for the similarity measure~\cite{Jegou-ECCV12}.
Additionnaly, they advocate the joint dimensionality reduction with multiple vocabularies to reduce the quantization artifacts underlying BOW and VLAD. % created using similar visual words vocabularies.
These vocabularies are created by using different initializations for the k-means algorithm, which may produce relatively highly correlated vocabularies.
%, and there exists a high probability that two local descriptors will be assigned to the same visual word in multiple similar vocabularies. 

In this paper, we propose to reduce the redundancy of the joint vocabulary representation (before the joint dimensionality reduction) by varying parameters of the local feature descriptors prior to the k-means quantization.
In particular, we propose: (i) different sizes of measurement regions for local description, (ii) different power-law normalizations of local feature descriptors, and (iii) different linear projections (PCA learned) to reduce the dimensionality of local descriptors. In this way, created vocabularies will be more complementary and joint dimensionality reduction of concatenated BOW vectors originating from several vocabularies will carry more information. Even though the proposed approaches are simple, we show that they provide significant boosts to retrieval performance with no memory or computational overhead at the query time.

%All proposed methods will be followed by evaluations on well known datasets. 

\paragraph{Related work.}%
% \subsubsection{Related work.}
This paper can be seen as an extension of~\cite{Jegou-ECCV12}, details of which are given later in Section~\ref{sec:baseline}. A number of papers report results with short descriptors obtained by PCA dimensionality reduction. In~\cite{Jegou-PAMI12} and~\cite{Perronnin-CVPR10}, aggregated descriptors (VLAD and Fisher vector respectively) are used followed by PCA to produce low dimensional image descriptors. In a paper about VLAD~\cite{Arandjelovic-CVPR13}, authors propose a method for adaptation of the vocabulary built on an independent dataset (adapt) and intra-normalization (innorm) method that $L_2$~normalizes all VLAD components independently, which suppresses the burstiness effect~\cite{Jegou-CVPR09}. In~\cite{Jegou-CVPR14}, a `democratic' weighted aggregation method for burstiness supression is introduced.
In this paper, we compare results of all the aforementioned methods using low dimensional descriptors $D'=128$.\\

The rest of the paper is organized as follows: Section~\ref{sec:background_and_baseline} gives a brief overview of several methods: bag-of-words\linebreak (BOW), efficient PCA dimensionality reduction of high dimensional vectors, and baseline retrieval with multiple vocabularies. Used datasets and evaluation protocols are established in Section~\ref{sec:datasets}. Section~\ref{sec:our_approach} introduces novel methods for joint dimensionality reduction of multiple vocabularies and presents extensive experimental evaluations. Main conclusions are given in Section~\ref{sec:conclusion}.

\section{Background and baseline}
\label{sec:background_and_baseline}

This section gives a short overview of the background of bag-of-words based image retrieval and the method used in~\cite{Jegou-ECCV12}. Key steps and ideas are discussed in higher detail to help understanding of the paper.

\subsection{Bag-of-words (BOW) image representation}
\label{sec:bow}

First efficient image retrieval based on BOW image representation was proposed by Sivic and Zisserman~\cite{Sivic-ICCV03}. They use local descriptors extracted in an image in order to construct a high-dimensional global descriptor. This procedure follows four basic steps:

\begin{enumerate}

\item For each image in the dataset, regions of interest are detected~\cite{Mikolajczyk-IJCV04,Matas-BMVC02} and described by an invariant descriptor which is $d$-dimensional. In this work we use the multi-scale Hessian-Affine~\cite{Perdoch-CVPR09} and MSER~\cite{Matas-BMVC02} detectors, followed by SIFT~\cite{Lowe-IJCV04} or RootSIFT~\cite{Arandjelovic-CVPR12} descriptors. The rotation of the descriptor is either determined by the detected dominant orientation~\cite{Lowe-IJCV04}, or by the gravity vector assumption~\cite{Perdoch-CVPR09}.
The descriptors are extracted from different sizes of measurement regions~\cite{Matas-BMVC02}, as described in detail in Section \ref{sec:our_approach}. 

\item Descriptors extracted from the training (independent) dataset (see Section \ref{sec:datasets}) are clustered into $k$ clusters using the k-means algorithm, which creates a visual vocabulary. 
%Descriptors from the dataset used for searching are quantized to generated visual vocabulary, producing visual words.

\item For each image in the dataset, a histogram of occurrences of visual words is computed. Different weighting schemes can be used, the most popular is inverse document frequency (\textit{idf}), which generates a $D$ dimensional BOW vector ($D = k$).

\item All resulting vectors are $L_2$~normalized, as suggested in~\cite{Sivic-ICCV03}, producing final global image representations used for searching.

\end{enumerate}

%\subsection{PCA reduction of multiple vocabularies}
%\label{sec:pca_main}
%
%In this section we overview methods for dimensionality reduction of high dimensional image representations and joint reduction of image representations created by multiple vocabularies. We also propose novel methods and show that they make significant improvements in the retrieval performance.

\subsection{Efficient PCA of high dimensional vectors}
\label{sec:pca}

In most of the cases BOW image representations have very high number of dimensions ($D$ can take values up to 64 million~\cite{Mikulik-IJCV12}). In these cases the standard PCA method (reducing $D$ to $D'$) computing the full covariance matrix is not efficient. The dual gram method (see Paragraph 12.1.4 in~\cite{Bishop06}) can be used to learn the first $D'$ eigenvectors and eigenvalues. Instead of computing the $D\times D$ covariance matrix $\boldsymbol{C}$, the dual gram method computes the $n\times n$ matrix $\boldsymbol{Y}^{T}\boldsymbol{Y}$, where $\boldsymbol{Y}$ is a set of vectors used for learning, and $n$ is the number of vectors in the set $\boldsymbol{Y}$. Eigenvalue decomposition is performed using the Arnoldi algorithm, which iteratively computes the $D'$ desired eigenvectors corresponding to the largest eigenvalues. This method is more efficient than the standard covariance matrix method if the number of vectors $n$ of the training set is smaller than the number of vector dimensions $D$, which is usually the case in the BOW approach.

Jegou and Chum~\cite{Jegou-ECCV12} analyze the effects of PCA dimensionality reduction on the BOW and VLAD vectors. They show that even though PCA successfully deals with the problem of negative evidence (higher importance of jointly missing visual words in compared BOW vectors), it ignores the problem of co-occurrences (co-occurences lead to over-count some visual patterns when comparing two image vector representation, see~\cite{Chum-CVPR10}). In order to tackle the aforementioned problem, they propose performing a whitening operation, similar to the one done in independent component analysis~\cite{Comon94} (implicitly performed by the Mahalanobis distance), jointly with the PCA. In our experiments we will use dimensionality reduction from $D$ to $D'$ components, as done in~\cite{Jegou-ECCV12}:

\begin{enumerate}

\item Every image vector $v = (v_1,\ldots, v_D)$ is post-processed using power-law normalization~\cite{Perronnin-CVPR10}: $v_i := |v_i|^{\beta}\times sign(v_i)$, with $0\leq \beta < 1$ as a fixed constant. Vector $v$ is $L_2$~normalized after processing. It has been shown~\cite{Jegou-PAMI12} that this simple procedure reduces the impact of multiple matches and visual bursts~\cite{Jegou-CVPR09}. In all our experiments $\beta = 0.5$, denoted as signed square rooting (SSR).

\item  First $D'$ eigenvectors of matrix $\boldsymbol{C}$ are learned using power-law normalized training vectors $\boldsymbol{Y} = [Y_1|\ldots |Y_n]$,  corresponding to the largest $D'$ eigenvalues $\lambda_{1},\ldots,\lambda_{D'}$.

\item Every power-law normalized image descriptor used for searching $X$ is PCA-projected and truncated, and at the same time whitened and re-normalized to a new vector $\hat{X}$ that is the final short vector representation with dimensionality $D'$:
\begin{equation}
\label{eq:pca_white}
\hat{X} = \frac
{\text{ diag}(\lambda_{1}^{-\frac{1}{2}},\ldots,\lambda_{D'}^{-\frac{1}{2}})\boldsymbol{P}^TX}
{\left \| \text{ diag}(\lambda_{1}^{-\frac{1}{2}},\ldots,\lambda_{D'}^{-\frac{1}{2}})\boldsymbol{P}^TX \right \|},
\end{equation}
where the $D\times D'$ matrix $\boldsymbol{P}$ is formed by the largest eigenvectors calculated in the previous step. Comparing two vectors after this dimensionality reduction with the Euclidian distance is now similar to using a Mahalanobis distance. It has been argued that the re-normalization step is critical for a better comparison metric, see~\cite{Jegou-ECCV12}.

\end{enumerate}

In order to compare results in a fair manner, we will use $D'=128$ dimensions for all our experiments following the trend of previous research in short image representations.

\subsection{The baseline method}
\label{sec:baseline}

\begin{figure*}\centering
\includegraphics[trim = 20mm 0mm 20mm 0mm, width=\linewidth]{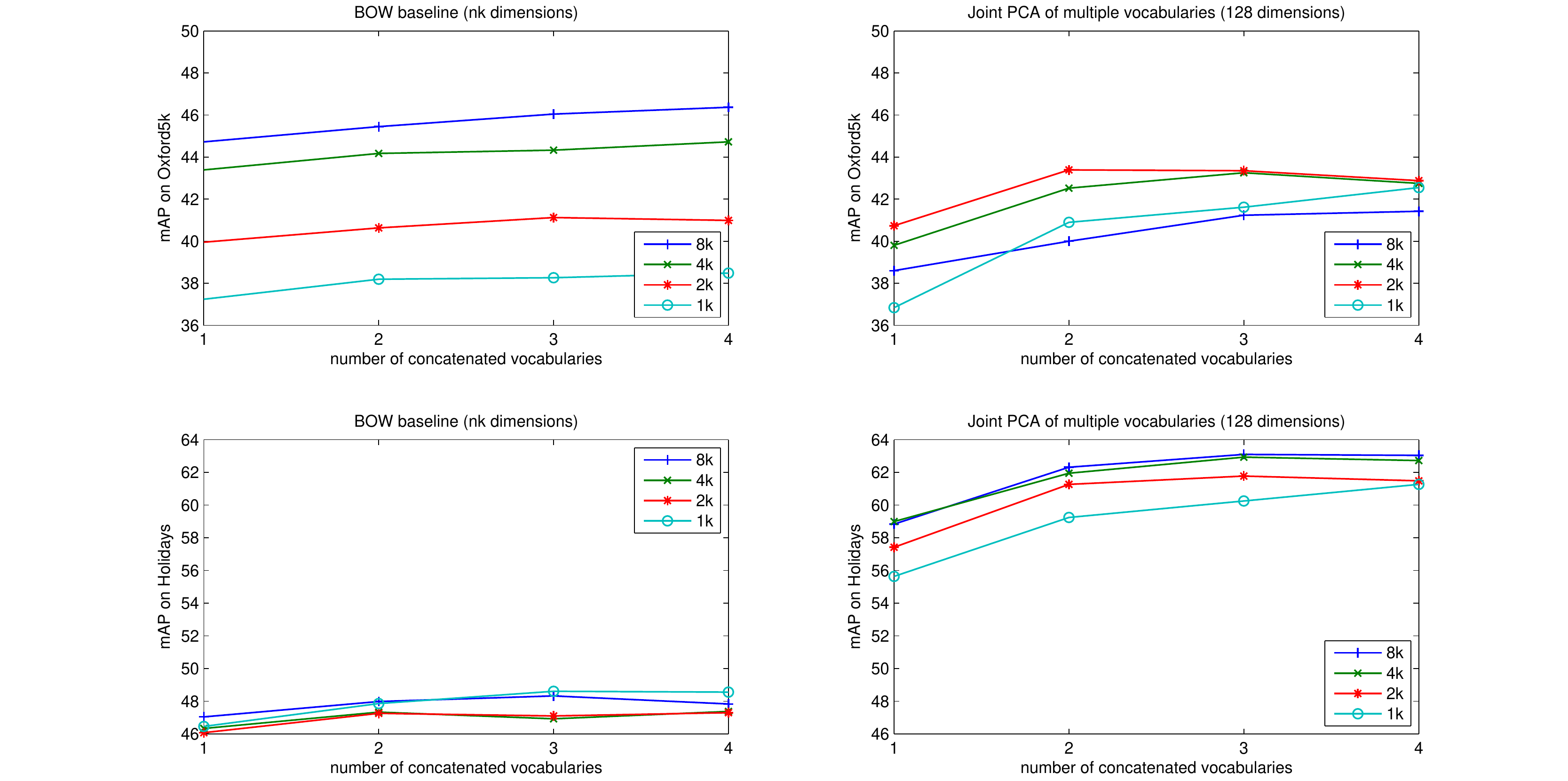}

\caption{\textbf{Baseline methods:} Left plots show mAP performance on Oxford5k (upper plot) and Holidays (lower plot) after straightforward concatenation of BOW vectors (no PCA dimensionality reduction performed) generated using multiple vocabularies. Note that dimensionality of BOW grows linearly with every new concatenation. Right plots present mAP performance on Oxford5k and Holidays after joint PCA dimensionality reduction of concatenated BOW representations to a $D'=128$ dimensional vector.}

\label{fig:related}
\end{figure*}

This paper builds upon the work~\cite{Jegou-ECCV12}, which is briefly reviewed in this section.
%
%There have been attempts of improving retrieval performance by using multiple vocabularies. It was shown that using multiple quantizations is a good way of dealing with quantization effects. First tries include hierarchical quantization method~\cite{Nister-CVPR06}, where vocabularies of different sizes were considered. Bad side of this approach is that size of the BOW vector grows with every vocabulary used. Other approach was proposed in~\cite{Jegou-PAMI10}, but it includes storing and querying more than one inverted file, increasing the memory requirements which are already very high. An example of straightforward concatenation of BOW vectors generated on multiple vocabularies can be seen in Figure~\ref{fig:related} (left plot). Performance improves slightly, but at the cost of linear increase in memory space. 
%
%As pointed out in~\cite{Jegou-ECCV12}, previous methods don't consider the relationship between vocabularies. Created from the same data, vocabularies are highly correlated, so there exists a high probability of two vectors being assigned to the same visual word in two different vocabularies. That is why,
%
In~\cite{Jegou-ECCV12}, a joint dimensionality reduction of multiple vocabularies is proposed. Image representation vectors are separately SSR normalized for each vocabulary, concatenated and then jointly PCA-reduced and whitened as explained in the Section~\ref{sec:pca}. The \textit{idf} term is ignored, and it is noted that the influence is limited when used with multiple vocabularies. Results of this method are shown in Figure~\ref{fig:related} (right plots). Comparing to the straightforward concatenation (Figure~\ref{fig:related}, left plots) where the results do not noticeably improve after adding multiple vocabularies, it can be noticed that an improvement in performance is achieved even when keeping low memory requirements by using PCA dimensionality reduction. However, for some vocabularies (i.e. $k=2$k), performance is dropping after only few vocabularies used.

%Multiple vocabularies are created using different random initialization of k-means algorithm and resulting clusters can be very similar (extremely highly correlated). Reducing dimensionality of BOW vectors created on mentioned vocabularies will partially solve this problem. To additionally alleviate mentioned problem, we propose several methods that boost results of PCA-reduced BOW vectors getting significant performance improvements even when combining small vocabularies, exceeding the state-of-the-art methods on short vector image representation.

% ******************************* Datasets **********************

\section{Datasets and evaluation}
\label{sec:datasets}

Results of our methods are evaluated on the datasets~\cite{Philbin-CVPR07,Philbin-CVPR08,Jegou-ECCV08} that are widely used in the image retrieval area. Also, we compare our results with other approaches evaluated on the same datasets.

\paragraph{Oxford5k~\cite{Philbin-CVPR07} and Paris6k~\cite{Philbin-CVPR08}:}%
% \subsubsection{Oxford5k~\cite{Philbin-CVPR07} and Paris6k~\cite{Philbin-CVPR08}:}
%
Both datasets contain a set of images (5062 for Oxford and 6300 for Paris) having 11 different landmarks together with distractors, downloaded from Flickr by searching for tags of popular landmarks. For each of the 11 landmarks there are 5 different query regions defined by a bounding box, meaning that there are 55 different query regions per dataset. The performance is reported as mean average precision (mAP), see~\cite{Philbin-CVPR07} for more details. In our experiments we use Paris6k as a training dataset in order to learn the visual vocabulary and projections of PCA dimensionality reduction. When evaluating our methods on Oxford5k, we always use the data learned on Paris6k.

\paragraph{Oxford105k~\cite{Philbin-CVPR07}:}% 
This dataset is the combination of Oxford5k dataset and 99782 negative images crawled from Flickr using 145 most popular tags. This dataset is used to evaluate the search performance (reported as mAP) on a large scale. Paris6k is used as a training dataset for Oxford105k.

\paragraph{Holidays~\cite{Jegou-ECCV08}:}% 
% \subsubsection{Holidays~\cite{Jegou-ECCV08}:} 
%
This dataset is a selection of personal holidays photos (1491 images) from INRIA, including a large variety of scene types (natural, man-made, water and fire effects, etc.). A sample of 500 images from the whole dataset is selected for query purposes~\cite{Jegou-ECCV08}. The performance is reported as mAP, like for Oxford5k and Oxford105k, after excluding the query image from the results. As a training dataset for vocabulary construction and image representation level PCA learning we use Paris6k dataset in all experiments.

% ******************************* Multiple Vocabularies **********************

\section{Sources of multiple codebooks}
\label{sec:our_approach}

We propose combining multiple vocabularies that are differing not just in random initialization of clustering procedure, but also in the data used for clustering. The feature data are alternated in the process of local features description.
This process is not trying to synthesize appearance deformations, but rather varying certain design choices in the pipeline of feature description, such as the relative size of the measurement region.  
Vocabularies created in this manner will contain less redundancy. This is combined with joint PCA dimensionality reduction (as described in Sections~\ref{sec:pca} and \ref{sec:baseline}) in order to produce short-vector image representations that are used for searching the most similar images in the dataset.

Quantization complexity for all vocabularies used in experiments is given in Table~\ref{tab:complexity}. As stated in~\cite{Jegou-ECCV12}, time necessary to quantize 2000 local descriptors of a query image, for four $k=8$k vocabularies, on 12 cores is 0.45s, using a multi-threaded exhaustive search implementation. Timings are proportional to the vocabulary size, i.e., to the number in the right column of Table~\ref{tab:complexity}. 

\setlength{\tabcolsep}{10pt}
\begin{table}
\centering
% \normalsize

\caption{\textbf{Complexity of vocabularies used throughout the experiments:} Complexity is given as a number of vector comparisons per local descriptor during the construction of the final BOW image representation.}

\begin{tabular}{lr}
\\
\hline
\textbf{Vocabulary} & \textbf{Complexity}\\
\hline
\hline
8k  & 8192\\
4k  & 4096\\
2k  & 2048\\
1k  & 1024\\
\hline
4k+2k+\ldots+128  & 8064\\
2k+1k+\ldots+128  & 3968\\
1k+512+256+128    & 1920\\
512+256+128       & 896\\
\hline
\end{tabular}%\\[\baselineskip]

\label{tab:complexity}

\end{table}

\paragraph{Multiple measurement regions.}%
% \subsubsection{Multiple measurement regions.} 
%

\begin{figure*} \centering

\begin{minipage}[c]{0.25\linewidth} \centering
\includegraphics[height=3cm]{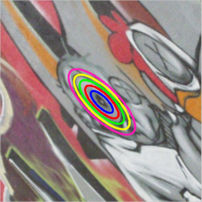}
\end{minipage}
\begin{minipage}[c]{0.74\linewidth} \centering
\setlength{\tabcolsep}{2pt}
\begin{tabular}{ccccc}
\includegraphics[height=2.2cm]{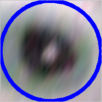} &
\includegraphics[height=2.2cm]{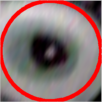} &
\includegraphics[height=2.2cm]{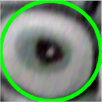} &
\includegraphics[height=2.2cm]{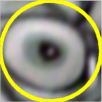} &
\includegraphics[height=2.2cm]{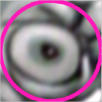}
\\
$0.5{\times}r$  &
$0.75{\times}r$ &
$1{\times}r$    &
$1.25{\times}r$ &
$1.5{\times}r$
\end{tabular}
\end{minipage}

\vspace{5pt}

\begin{minipage}[c]{0.25\linewidth} \centering
\includegraphics[height=3cm]{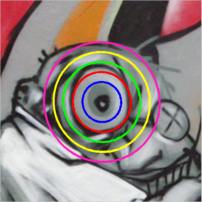}
\end{minipage}
\begin{minipage}[c]{0.74\linewidth} \centering
\setlength{\tabcolsep}{2pt}
\begin{tabular}{ccccc}
\includegraphics[height=2.2cm]{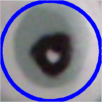} &
\includegraphics[height=2.2cm]{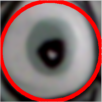} &
\includegraphics[height=2.2cm]{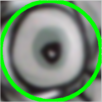} &
\includegraphics[height=2.2cm]{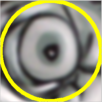} &
\includegraphics[height=2.2cm]{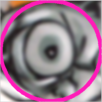}
\\
$0.5{\times}r$  &
$0.75{\times}r$ &
$1{\times}r$    &
$1.25{\times}r$ &
$1.5{\times}r$
\end{tabular}
\end{minipage}

\vspace{2mm}
\caption{\textbf{Multiple measurement regions (mMeasReg):} A corresponding feature is detected in two images (left). Multiple measurement regions for a single detected feature are shown in each row. The normalized patches (right) show different image content described by the respective descriptor.} \label{fig:multiMR}

\end{figure*}

An affine invariant descriptor of an affine covariant region can be extracted from any affine covariant constructed measurement region~\cite{Matas-BMVC02}. As an example of a measurement region that is, in general, of a different shape than the detected region, is an ellipse fitted to the regions, as proposed by~\cite{Tuytelaars-BMVC00} and also used for MSERs~\cite{Matas-BMVC02}. An important parameter is the relative scale of the measurement region with respect to the scale of the detected region. Since the output of the detector is designed to be repeatable, it is usually not discriminative. To increase the disriminability of the descriptor, it is commonly extracted from area larger than the detected region. In case of~\cite{Perdoch-CVPR09}, the relative change in the radius is $r=3\sqrt{3}$. The larger the region, the higher discriminability of the descriptor, as long as the measurement region covers a close-to-planar surface. On the other hand, larger image patches have higher chance of hitting depth discontinuities and thus being corrupted. An example of multiple measurement regions is shown in Figure~\ref{fig:multiMR}. To take the best of this trade off, we propose to construct multiple vocabularies over descriptors extracted at multiple relative scales of the measurement regions. Including lower scales leverages the disadvantages of large measurement regions, while joint dimensionality reduction eliminates the dependencies between the representations. 

\begin{figure*}\centering
\includegraphics[trim = 20mm 0mm 20mm 0mm, width=\linewidth]{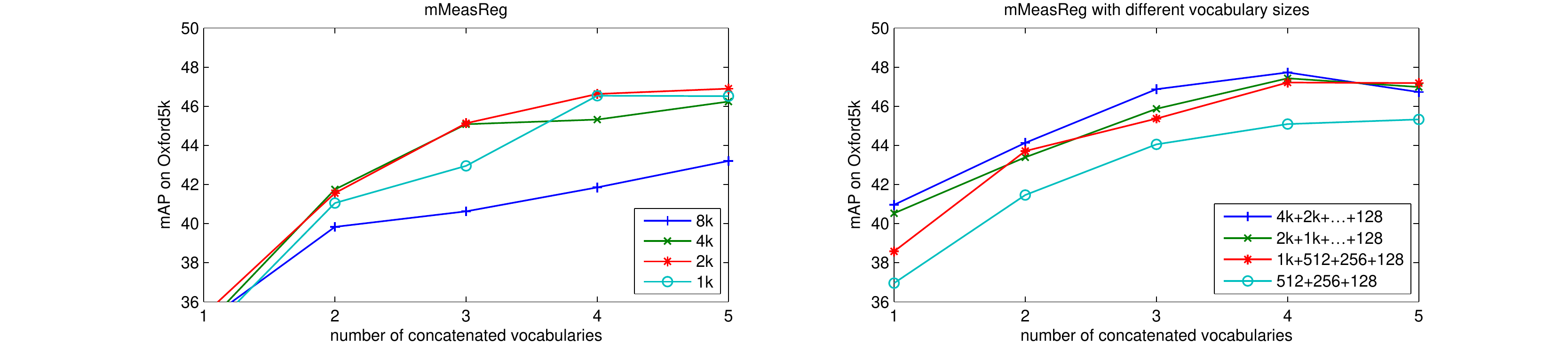}

\caption{\textbf{Multiple measurement regions (mMeasReg):} mAP performance improvement on Oxford5k after PCA reduction to $D'=128$ of concatenated BOW vectors produced on vocabularies created using SIFT descriptors with different measurement regions: $0.5{\times}r,\: 0.75{\times}r,\: 1{\times}r,\: 1.25{\times}r,\: 1.5{\times}r$.}

\label{fig:mes_reg}
\end{figure*}

%
%computed over its affine-normalized local neighborhood. In other words, patches around interest points are considered with a predefined size called measurement region. Commonly used measurement region has the radius $r=3\sqrt{3}s$, where $s=(det(\boldsymbol{A})^{-1/2}$ is the scale of the interest point and $\boldsymbol{A}$ is the transformation mapping points on the ellipse to points on a unit circle, see~\cite{Perdoch-CVPR09} for more details. 
%
We consider using different sizes of measurement regions: $0.5\times r,\: 0.75\times r,\: 1\times r,\: 1.25\times r,\: 1.5\times r$; creating slightly different SIFT descriptors used to learn every vocabulary. Implementation is very simple and during online stage the computation has to be done only for the features from query image region. Though simple, this method provides significant improvement even when concatenating vocabularies of small sizes (i.e. $k=2\text{k}$ and $k=1\text{k}$), see Figure~\ref{fig:mes_reg} (left plot). We also explore the use of vocabularies with different sizes. All BOW vectors in this case are weighted proportionally to the logarithm of their vocabulary size~\cite{Jegou-ECCV12}. In each step we concatenate a new bundle of vocabularies with multiple sizes, calculated with a different measurement region. We notice improvement when using multiple vocabulary sizes as well, see Figure~\ref{fig:mes_reg} (right plot). For presentation of results on both plots in Figure~\ref{fig:mes_reg}, in every step we are adding a different vocabulary created on SIFT vectors with measurement regions in predefined order: $0.5\times r,\: 0.75\times r,\: 1\times r,\: 1.25\times r,\: 1.5\times r$. This approach is denoted as mMeasReg.

\paragraph{Multiple power-law normalized SIFT descriptors.}% 
% \subsubsection{Multiple power-law normalized SIFT descriptors.} 
%
% \linebreak 
SIFT descriptors~\cite{Lowe-IJCV04} were the popular choice in most of the image retrieval systems for a long time. Arandjelovic \etal~\cite{Arandjelovic-CVPR12} show that using a Hellinger kernel instead of standard Euclidian distance to measure the similarity between SIFT descriptors leads to a noticeable performance boost in retrieval system. The kernel is implemented by simply square rooting every component of SIFT descriptor. Using Euclidian distance on these new RootSIFT descriptors will give the same result as using Hellinger kernel on the original SIFT descriptors. 
In general, a power-law normalization~\cite{Perronnin-CVPR10} with any power $0 \leq \beta \leq 1$ can be applied to the descriptors ($\beta = 0.5$ resulting in RootSIFT~\cite{Arandjelovic-CVPR12}). Voronoi cells constructed in power-law normalized descriptor spaces can be seen as non-linear hyper-surfaces separating the features in the original (SIFT) descriptor space. Concatenation of such feature space partitionings reduces the redundant information.  

\begin{figure*}[t!]\centering
\includegraphics[trim = 20mm 0mm 20mm 0mm, width=\linewidth]{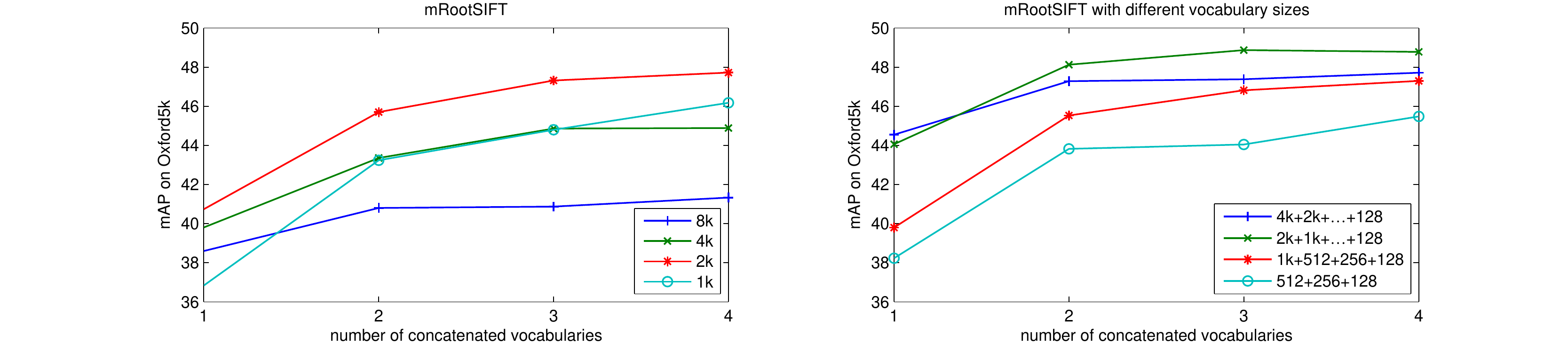}

\caption{\textbf{Multiple power-law normalized SIFT descriptors (mRootSIFT):} mAP performance improvement on Oxford5k after PCA reduction to $D'=128$ of concatenated BOW vectors produced on vocabularies created using multiple local feature descriptors: SIFT, $\text{SIFT}^{0.4}$, $\text{SIFT}^{0.5}$, $\text{SIFT}^{0.6}$.}

\label{fig:SIFTs}
\end{figure*}

There is no additional memory required and the change can be done on-the-fly with virtually no additional computational cost using simple power operation. We consider building four different vocabularies using: SIFT and SIFT with every component to the power of 0.4, 0.5, 0.6 (denoted as $\text{SIFT}^{0.4}$, $\text{SIFT}^{0.5}$, $\text{SIFT}^{0.6}$ respectively). Concatenation is done on single vocabularies (Figure~\ref{fig:SIFTs}, left plot) and on a bundle of vocabularies with different sizes (Figure~\ref{fig:SIFTs}, right plot). Adding all SIFT modifications to the process of vocabulary creation achieves noticeable improvement of retrieval performance in the case of all vocabulary sizes. We denote this method as mRootSIFT.

Combining vocabularies of different SIFT exponents improves over combining different vocabularies of a single SIFT exponent. For example, for 4~$\times$ 2k vocabularies, the mAP on Oxford5k is $46.5$ for 4~$\times$ $\text{SIFT}^{0.5}$, and $47.7$ (Figure~\ref{fig:SIFTs} left) for exponent combination.

% \paragraph{Gravity vector.}

\paragraph{Multiple linear projections of SIFT descriptors.}%
% \subsubsection{Multiple linear projections of SIFT descriptors.}
%
In locality sensitive hashing (random) linear projections are commonly used to reduce the dimensionality of the space while preserving locality. The idea pursued in this part of the paper is to use linear projections on the feature descriptors (SIFTs) before the vocabulary construction via k-means. However, random projections do not reflect the structure of the descriptors, resulting in noisy descriptor space partitionings. We propose to use PCA learned linear projections of SIFTs, learned on different training sets or subsets. The projections learned this way account for the statistics given by the training sets and hence produce meaningful distances, while inserting different biases into the vocabulary construction.

\begin{figure*}[t!]\centering
\includegraphics[trim = 20mm 0mm 20mm 0mm, width=\linewidth]{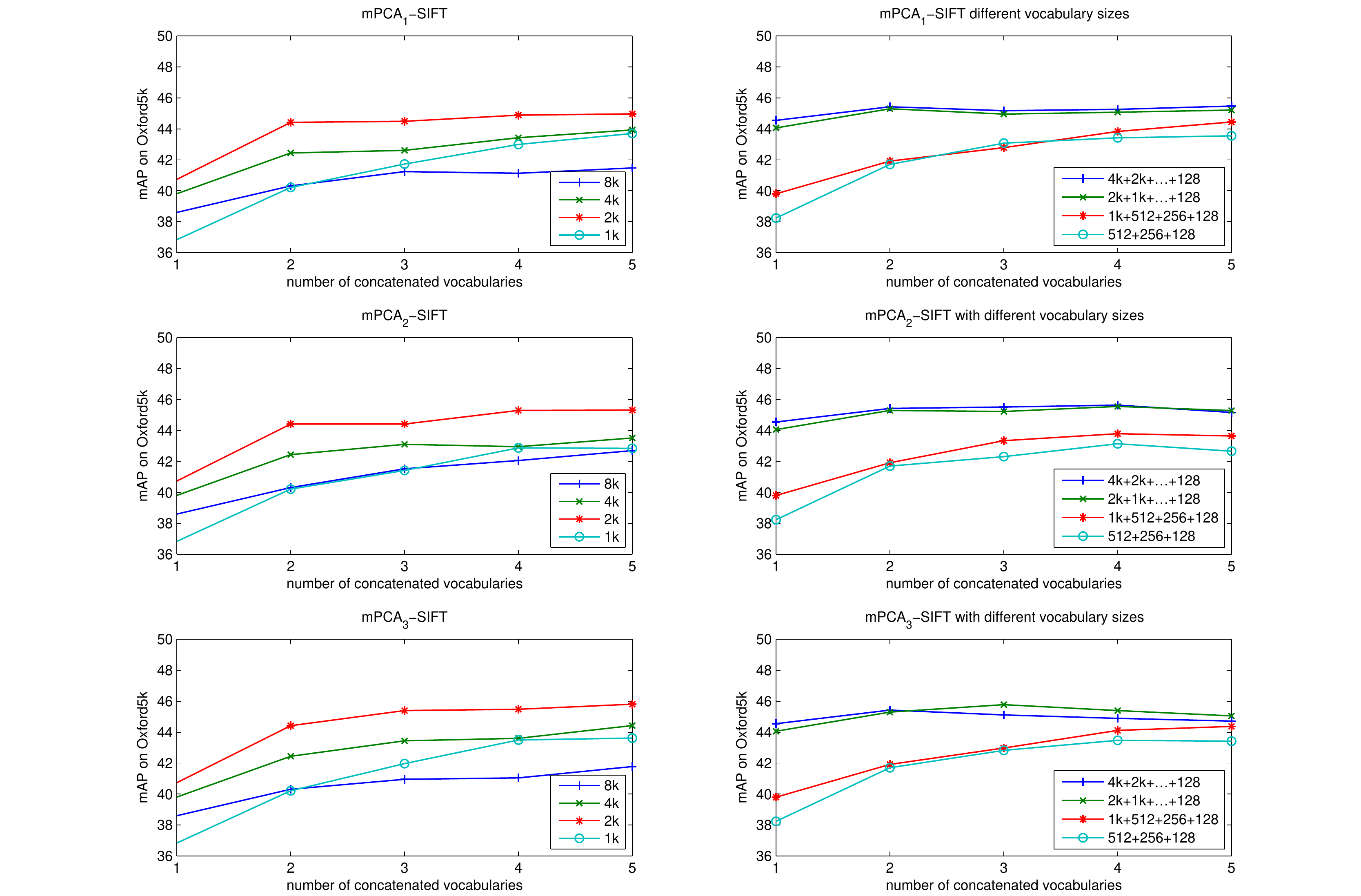}

\caption{\textbf{Multiple linear projections of SIFT descriptors (mPCA-SIFT):} mAP performance improvement on Oxford5k after PCA reduction to $D'=128$ of concatenated BOW vectors produced on vocabularies created using different PCA-reduced SIFT descriptors. For more details about all three presented methods see Section~\ref{sec:our_approach}.}

\label{fig:PCA}
\end{figure*}

%In the last experiment we consider doing standard PCA dimensionality reduction \cite{Bishop06} of local ($D=128$ dimensional) SIFT descriptors to $D'$ prior to creation of multiple vocabularies and aggregation of BOW vectors. 

The improvement is twofold: (i) increased performance measured by mAP, and (ii) shorter quantization time during query due to shorter local descriptors after dimensionality reduction. On the other side there is a small amount of storage required to save learned projection matrices for every vocabulary, which we reuse at query. We consider and evaluate three different approaches for learning the eigenvectors used to project SIFT vectors from $D$ to $D'$ dimensions:
\begin{enumerate}[nolistsep]

\item We learn eigenvectors on Paris6k dataset and reduce the dimension of SIFT descriptors to $D'=80,64,48,32$ in the respective order for every newly created vocabulary (m$\text{PCA}_1$-SIFT). Results of this experiment are shown in Figure~\ref{fig:PCA}, \nth{1} row.

\item We learn eigenvectors on different datasets: Paris6k, Holidays, University of Kentucky benchmark (UKB), PASCAL VOC'07 training in the respective order for every newly created vocabulary (m$\text{PCA}_2$-SIFT). Dimension of SIFT descriptors is reduced to $D'=80$ in all cases. For the mAP performance on Oxford5k, see Figure~\ref{fig:PCA}, \nth{2} row.

\item We learn eigenvectors on different datasets: Paris5k, Holidays, UKB, PASCAL VOC'07 training and reduce the dimension of SIFT descriptors differently for each dataset ($D'=80,64,48,32$ respectively) creating different vocabularies (m$\text{PCA}_3$-SIFT). Performance is presented in Figure~\ref{fig:PCA}, \nth{3} row.

\end{enumerate}
Note that first vocabulary in all three different approaches is produced using standard SIFT descriptors without PCA reduction. A new vocabulary is added in every step of the experiment having joint dimensionality reduction of 5 concatenated BOW vectors in the end.

% \bigskip
\paragraph{Multiple feature detectors.}% 
% \subsubsection{Multiple feature detectors.} 
%
In the Video Google approach~\cite{Sivic-ICCV03} the authors combine vocabularies created from two different feature types.
%In the Video Google approach~\cite{Sivic-ICCV03}, the authors detect two types of features and create separate vocabularies combining them into one vocabulary at the end. 
In this paper we attempt to combine Hessian-Affine~\cite{Perdoch-CVPR09} and MSER~\cite{Matas-BMVC02} detectors. Even though straightforward concatenation of BOW vectors created on $k=8$k vocabularies ($48.7$ mAP on Oxford5k) gives improvement over using single BOW representations with Hessian-Affine ($44.7$) and MSER ($40.1$) features, after joint PCA reduction there is a decrease of performance when combining features ($37.0$ mAP on Oxford5k) compared to only doing PCA reduction on a single Hessian-Affine vocabulary ($38.6$), and an increase in performance when compared to PCA-reduced BOW vectors built on a single MSER vocabulary ($24.4$). Similar conclusions are made when combining smaller vocabulary sizes, i.e., there is always a drop in performance when comparing PCA reduction on a single vocabulary with Hessian-Affine features and PCA on combined vocabularies with Hessian-Affine and MSER features; mAP drop: from $39.8$ to $39.1$, from $40.7$ to $38.7$, from $36.8$ to $35.1$ for $k=4$k, $2$k, $1$k respectively. We also experimented with combining Harris-Affine~\cite{Mikolajczyk-IJCV05} with Hessian-Affine features in the same manner as with MSER, but the improvement is not significant. PCA reduction of a single $k=8$k vocabulary on Hessian-Affine yields $38.6$ mAP on Oxford5k while joint PCA after adding a vocabulary of the same size built on Harris-Affine improves mAP to $39.0$, which is smaller improvement than using two vocabularies built on Hessian-Affine features with different randomization ($40.0$ mAP).

\begin{figure}[b!]\centering
\includegraphics[width=.9\columnwidth]{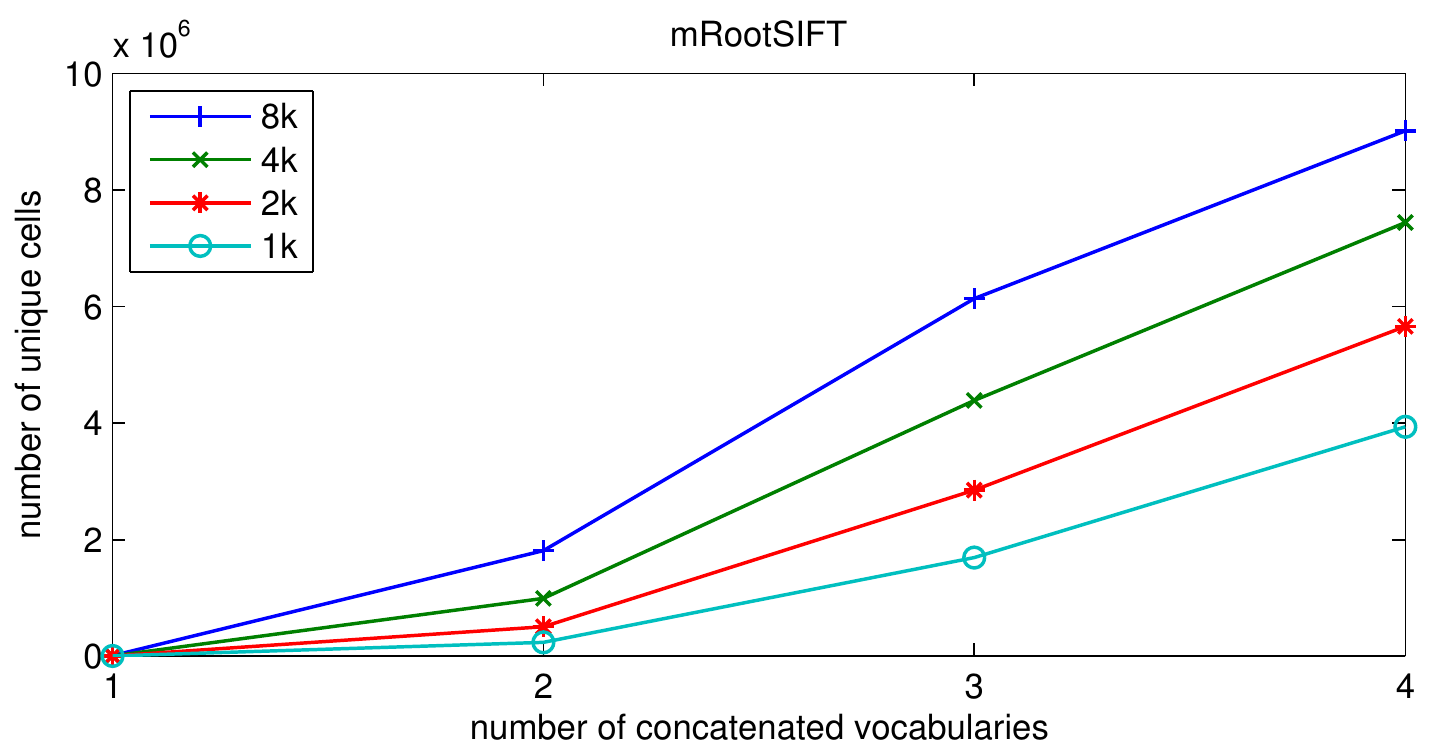}
\caption{\textbf{Number of unique assignments (vocabulary cells) for Oxford5k dataset when combining vocabularies built on multiple power-law normalized SIFT descriptors (mRootSIFT):} SIFT, $\text{SIFT}^{0.4}$, $\text{SIFT}^{0.5}$, $\text{SIFT}^{0.6}$.}
\label{fig:unq}
\end{figure}

\paragraph{Discussion.}%
In order to better understand the impact of using multiple vocabularies we count the number of unique assignments in the product vocabulary. It corresponds to the number of non-empty cells of the descriptor space generated by all vocabularies simultaneously. The maximum possible number of unique assignments is equal to the product of number of clusters (cells) of all joint vocabularies. The number is related to the precision of reconstruction of each feature descriptor from its visual word assignments. For combination of vocabularies with different SIFT exponents (mRootSIFT) the number of unique assignments for Oxford5k dataset is shown in Figure~\ref{fig:unq}. The plots are similar for all vocabulary combinations.

\setlength{\tabcolsep}{8pt}
\begin{table*}
\centering
% \normalsize

\caption{\textbf{Comparison with the state-of-the-art on short vector image representation ($D'=128$):} Results in the first section of the table are mostly obtained from the paper~\cite{Jegou-PAMI12}, except for the recent method on triangulation embedding and democratic aggregation with rotation and normalization ($\phi_\Delta$+$\psi_\text{d}$+$\text{RN}$) proposed in~\cite{Jegou-CVPR14}. In the second section we present results from methods that are using joint PCA and whitening of high dimensional vectors as we do. Results marked with * are obtained after our reimplementation of the methods using feature detector and descriptor as described in Section~\ref{sec:bow} and Paris6k as a learning dataset. In the last section of the table we present results of our methods. All methods are described in detail in Section~\ref{sec:our_approach}.}

% \resizebox{\linewidth}{!}{%
\begin{tabular}{llccc}
\\
\hline
\textbf{Method} & \textbf{Vocabulary} & \textbf{Oxford5k} & \textbf{Oxford105k} & \textbf{Holidays}\\
\hline
\hline
GIST~\cite{Oliva-IJCV01}     & N/A & $-$ & $-$ & $36.5$\\
BOW~\cite{Sivic-ICCV03}      & $k{=}20\text{k}$ & $19.4$ & $-$ & $45.2$\\
Improved Fisher~\cite{Perronnin-CVPR10} & $k{=}64$ & $30.1$ & $-$ & $56.5$\\
VLAD~\cite{Jegou-CVPR10} & $k{=}64$ & $-$ & $-$ & $51.0$\\
VLAD+SSR~\cite{Jegou-PAMI12} & $k{=}64$ & $28.7$ & $-$ & $55.7$\\
$\phi_\Delta$+$\psi_\text{d}$+RN~\cite{Jegou-CVPR14} & $k{=}16$ & $43.3$ & $35.3$ & $61.7$\\
\hline
mVocab/BOW~\cite{Jegou-ECCV12} & $k{=}4{\times}8\text{k}$ & $41.3{/}41.4^{*}$ & $-{/}33.2^{*}$ & $56.7{/}63.0^{*}$\\
mVocab/BOW~\cite{Jegou-ECCV12} & $k{=}2{\times}(32\text{k}{+}\ldots{+}128)$ & $-{/}42.9^{*}$ & $-{/}35.1^{*}$ & $60.0/64.5^{*}$\\
mVocab/VLAD~\cite{Jegou-ECCV12} & $k{=}4{\times}256$ & $-$ & $-$ & $61.4$\\
mVocab/VLAD+adapt+innorm~\cite{Arandjelovic-CVPR13} & $k{=}4{\times}256$ & $44.8$ & $37.4$ & $62.5$\\
\hline
mMeasReg/mVocab/BOW & $k{=}5{\times}2\text{k}$ & $46.9$ & $38.9$ & $66.9$\\
mMeasReg/mVocab/BOW & $k{=}4{\times}(4\text{k}{+}\ldots{+}128)$ & $47.7$ & $39.2$ & $67.3$\\
mRootSIFT/mVocab/BOW & $k{=}4{\times}2\text{k}$ & $47.7$ & $39.8$ & $64.3$\\
mRootSIFT/mVocab/BOW & $k{=}4{\times}(2\text{k}{+}\ldots{+}128)$ & $48.8$ & $41.4$ & $65.6$\\
m$\text{PCA}_3$-SIFT/mVocab/BOW & $k{=}5{\times}2\text{k}$ & $45.8$ & $38.1$ & $63.2$\\
m$\text{PCA}_1$-SIFT/mVocab/BOW & $k{=}5{\times}(4\text{k}{+}\ldots{+}128)$ & $45.5$ & $37.8$ & $64.6$\\
\hline
\end{tabular}%\\[\baselineskip] 
% }
\label{tab:mAP}
\end{table*}

\subsection{Comparison with the state-of-the-art}

Comparison with the current methods dealing with short vector image representation is given in Table~\ref{tab:mAP}. Authors of the baseline approach on multiple vocabularies (mVocab) did not provide results for Oxford5k and Oxford105k datasets using all of their proposed methods, so we reimplemented and presented the corresponding results. Compared to their best method on Oxford5k that achieves $42.9$ mAP, our best method ($48.8$ mAP) obtains significant relative improvement of $13.8\%$. In fact, all our methods outperform mVocab baseline methods on Oxford5k by a noticeable margin, with an improvement of $6.1\%$ in the case of our worst performing method. When evaluating large-scale retrieval on Oxford105k dataset our methods again outperform the baseline method, relative improvement is $17.9\%$ for our best performing method, and $7.7\%$ for the worst performing one. In order to make a fair comparison when evaluating on Holidays dataset we again reimplemented the baseline approach, using Paris6k for learning the vocabularies and PCA projections (as we did in all our methods). In this case, the relative improvement is $4.3\%$ with our best method (from $64.5$ mAP to $67.3$ mAP). We also compare our methods to two recent state-of-the-art approaches on short representations~\cite{Arandjelovic-CVPR13,Jegou-CVPR14}. On Oxford5k and Oxford105k we improve as much as $8.9\%$ and $10.7\%$, respectively, compared to VLAD based approach~\cite{Arandjelovic-CVPR13}, and $12.7\%$ and $17.3\%$, respectively, compared to T-embedding based approach~\cite{Jegou-CVPR14}. On Holidays dataset relative improvement is $7.7\%$ compared to the former and $9.1\%$ compared to the latter. Note that the dataset used for learning of the meta-data for Holidays is different: we use Paris6k, while both \cite{Arandjelovic-CVPR13} and \cite{Jegou-CVPR14} are using an independent dataset comprising of 60k images downloaded from Flickr.

\section{Conclusions}
\label{sec:conclusion}

Methods for multiple vocabulary construction were studied and evaluated in this paper. Following~\cite{Jegou-ECCV12}, the concatenated BOW image representations from multiple vocabularies were subject to joint dimensionality reduction to 128D descriptors. We have experimentally shown that generating diverse multiple vocabularies has crucial impact on search performance. Each of the multiple vocabularies was learned on local feature descriptors obtained with varying parameter settings. That includes
feature descriptors extracted from measurement regions of different scales, different power-law normalizations of the SIFT descriptors, and applying different linear projections to feature descriptors prior to k-means quantization. 
The proposed vocabulary constructions improve performance over the baseline method~\cite{Jegou-ECCV12}, where only different initializations were used to produce multiple vocabularies. More importantly, the {\em all} proposed methods exceed the state-of-the-art results~\cite{Arandjelovic-CVPR13,Jegou-CVPR14} by a large margin. The choice of the optimal combination of vocabularies to combine still remains an open problem.

\clearpage

\vspace{2mm} \noindent {\bf Acknowledgements.} The authors were supported by MSMT LL1303 ERC-CZ and ERC VIAMASS no. 336054 grants.

% \section*{Acknowledgements}

{\small
\bibliographystyle{ieee}
\bibliography{multiple.bib}
}

%\flushend

\end{document}